\documentclass[10pt,twocolumn,letterpaper]{article}

\usepackage{cvpr}              %

\usepackage{graphicx}
\usepackage{amsmath}
\usepackage{amssymb}
\usepackage{booktabs}
\usepackage{float}
\usepackage{titling}
\usepackage{enumitem}
\usepackage{paralist}
\usepackage{multicol}
\usepackage[symbol]{footmisc}
\usepackage{xcolor}
\definecolor{xucongcolor}{rgb}{0.73725, 0.6588, 0.0705}

\usepackage{xspace}
\newcommand{\methodname}[0]{\textrm{EFE}\xspace}
\newcommand{\parahead}[1]{\vspace{1mm}\noindent\textbf{#1.}}

\usepackage{subcaption}
\usepackage{array}

\usepackage[pagebackref,breaklinks,colorlinks,hyperfootnotes=false]{hyperref}

\usepackage[capitalize]{cleveref}
\crefname{section}{Sec.}{Secs.}
\Crefname{section}{Section}{Sections}
\Crefname{table}{Table}{Tables}
\crefname{table}{Tab.}{Tabs.}

\def\cvprPaperID{6} %
\def\confName{GAZE}
\def\confYear{2023}

\title{\methodname: End-to-end Frame-to-Gaze Estimation}
\author{
Haldun Balim\textsuperscript{1} \hspace{3mm}
Seonwook Park\textsuperscript{2}\footnotemark[1] \hspace{3mm}
Xi Wang\textsuperscript{1}\footnotemark[1] \hspace{3mm}
Xucong Zhang\textsuperscript{3}\footnotemark[1] \hspace{3mm}
Otmar Hilliges\textsuperscript{1} \\[1mm]
\textsuperscript{1}Department of Computer Science, ETH Z\"urich \quad
\textsuperscript{2}Lunit Inc. \quad \\
\textsuperscript{3}Computer Vision Lab, Delft University of Technology
}

\begin{document}
\date{}

\twocolumn[{%
\renewcommand\twocolumn[1][]{#1}%
\maketitle
\begin{center}
    \centering
    \captionsetup{type=figure}  
    \vspace*{-4mm}
    \includegraphics[width=0.95\textwidth]{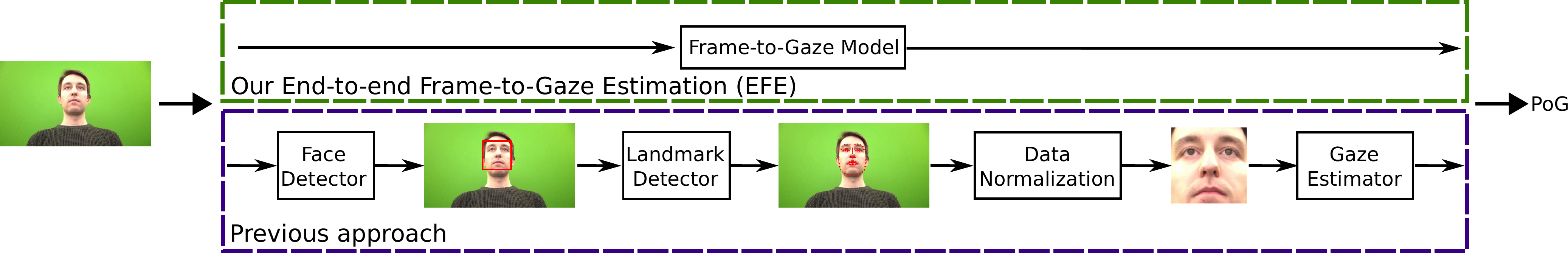}
    \captionof{figure}{\textbf{Comparison of the proposed \methodname method against conventional gaze estimation methods.} Our \textbf{E}nd-to-end \textbf{F}rame-to-Gaze \textbf{E}stimation approach (\methodname) is trained to predict eye gaze directly from the input camera frame. In contrast, most existing approaches rely on additional pre-processing modules. For example, face and facial landmark detection methods are used in ``data normalization'' to obtain eye or face patches, which are then used as inputs to gaze estimation models. Our approach, \methodname, demonstrates that it is possible to skip those steps while maintaining or improving performance.
        \label{fig:teaser}
    }
\end{center}
}]

\footnotetext[1]{These authors contributed equally to this work.}

\begin{abstract}
Despite the recent development of learning-based gaze estimation methods, most methods require one or more eye or face region crops as inputs and produce a gaze direction vector as output.
Cropping results in a higher resolution in the eye regions and having fewer confounding factors (such as clothing and hair) is believed to benefit the final model performance.
However, this eye/face patch cropping process is expensive, erroneous, and implementation-specific for different methods.
In this paper, we propose a frame-to-gaze network that directly predicts both 3D gaze origin and 3D gaze direction from the raw frame out of the camera without any face or eye cropping.
Our method demonstrates that direct gaze regression from the raw downscaled frame, from FHD/HD to VGA/HVGA resolution, is possible despite the challenges of having very few pixels in the eye region.
The proposed method achieves comparable results to state-of-the-art methods in Point-of-Gaze (PoG) estimation on three public gaze datasets: GazeCapture, MPIIFaceGaze, and EVE, and generalizes well to extreme camera view changes. 

\end{abstract}

\section{Introduction}

Remote webcam-based gaze estimation is a well-studied problem setting where images from a single user-facing and remotely-placed camera are used to estimate the gaze of a user.
Effective solutions to this problem can enable novel applications in gaze-contingent human-computer interaction~\cite{majaranta2014eye,biedert2010text}, adaptive user interfaces~\cite{feit2020detecting,kajan2016peeplist}, and crowd-sourced attention studies~\cite{papoutsaki2016webgazer}.
Unlike infrared-light devices~\cite{zhu2007novel}, webcam-based gaze estimation can allow for large operating distances~\cite{zhang2019evaluation}.
With the introduction of large in-the-wild datasets~\cite{Zhang2015CVPR,Krafka2016CVPR}, many learning-based convolutional neural network-based (CNN) approaches~\cite{zhang2022gazeonce, park2020towards, Zhang2015CVPR, wang2022contrastive} have been proposed, %
enabling gaze estimation from a single front-facing camera.

Learning-based remote gaze estimation methods typically take small cropped patches as input to predict the gaze direction. 
These inputs as well as gaze origin must be generated with pre-defined processes according to the facial landmarks. 
Specifically, inputs to these methods are either simply cropped images \cite{Krafka2016CVPR,huang2017tabletgaze} or image patches yielded via a process known as ``data normalization'' \cite{Sugano2014CVPR,Zhang2018ETRA,park2020towards}.
Simple cropping methods usually simply crop the eye or face according to the facial landmarks. Since it is cropping directly on the 2D image without consideration of the 3D head pose, it could result in different sizes and image ratios in the case of large head rotations. This can introduce unnecessary appearance variations for gaze estimator training, reducing performance~\cite{Zhang2018ETRA}.
With the simple cropped face and eye images, recent gaze estimators can perform cross-person gaze estimation by leveraging a large amount of training data from multiple subjects~\cite{Krafka2016CVPR}. 
To be able to directly regress to on-screen Point-of-Gaze (PoG) without gaze origin and gaze direction predictions, these methods have to assume that the camera plane is coplanar to the screen plane~\cite{Krafka2016CVPR,huang2017tabletgaze}. However, such a coplanarity assumption cannot be easily held for many application scenarios.

To eliminate the coplanar assumption, we could first calculate the gaze origin and then take the face/eye crop to predict the gaze direction as a two-step approach.
To obtain the gaze origin and face/eye crop, data normalization is proposed as a pre-processing step \cite{Sugano2014CVPR,Zhang2018ETRA}.
As shown in the bottom of Fig. \ref{fig:teaser}, it crops the eye/face patch out of the input camera frame according to the facial landmarks and estimates a 3D head pose by fitting a generic 3D face model, which is further used to yield the 3D gaze origin.
The gaze estimation method then only needs to output the gaze direction in the camera coordinate system.
By composing the predicted gaze direction with the gaze origin acquired at the data normalization step, the gaze ray can be constructed.
Note there is no joint optimization of gaze origin and gaze direction since the gaze origin is estimated separately from facial landmarks and explicit head pose estimation.
The ill-posed problem of 3D head translation estimation from a 2D image could introduce extra error in the depth estimation of the gaze origin.
Furthermore, the processes of data normalization are often implemented as an expensive offline procedure (see~\cite{Park2019ICCV,park2020towards} in particular).

Skipping the aforementioned pre-processing steps and directly taking the raw frame as input 
is a highly challenging task and has rarely been investigated in the literature due to small eye region and gaze original estimation.
Since CNNs typically accept small images around $224\times224$ to $512\times512$ pixels, we have to resize the raw input frame to be much smaller from HD (720p) or Full HD (1080p) resolution. It results in a much smaller region of interest (face/eye) than the cropped patches.
Without the facial landmark, the 3D location of the gaze origin needs to be accurately estimated from the monocular image which has not been investigated in previous works.

In this paper, we demonstrate that it is possible to train an ``\textbf{E}nd-to-end \textbf{F}rame-to-Gaze \textbf{E}stimation'' (\methodname) method without making the aforementioned coplanar assumption.
As shown in the top of Fig. \ref{fig:teaser}, our approach avoids the need for expensive ``data normalization'' and directly regresses a 6D gaze ray (3D origin and 3D direction) from the camera frame without cropping the face/eye, allowing the trained method to adapt to new camera-screen geometries. The PoG is calculated by the 3D origin and 3D gaze direction.
We observe that the gaze origin can be predicted accurately with a fully-convolutional U-Net-like architecture (shown in Fig. \ref{fig:overview}). The network predicts a 2D heatmap ($\mathbf{h}$), which captures the 2D location of the gaze origin inside of the frame, and a depth map ($\mathbf{d}$) that captures the distance to the gaze origin in the third dimension.
We take the bottleneck features from the U-Net-like architecture and pass them through a multi-layer perception (MLP) to predict the 3D gaze direction. 
With camera intrinsic and extrinsic parameters, we intersect the gaze ray with the known screen plane in a differentiable manner to yield PoG.
This architecture can be trained end-to-end and we evaluate it on three existing large datasets: EVE~\cite{park2020towards}, GazeCapture~\cite{Krafka2016CVPR}, and MPIIFaceGaze~\cite{Zhang2019TPAMI}. 
We show that it achieves comparable performance with competitive baselines using ``data normalized'' inputs on the EVE,
GazeCapture and MPIIFaceGaze datasets.

\section{Related Works}

\subsection{Remote Gaze Estimation from RGB}
Remote gaze estimation from RGB is a setting where a single RGB camera faces the user and no additional instruments such as IR light sources are used to make the gaze estimation problem more tractable.
Due to the challenges imposed by this setting, even early methods for gaze estimation tend to apply machine learning techniques~\cite{baluja1993non,sugano2008incremental,lu2014adaptive,sesma2012evaluation} but are limited to tackling person-specific gaze estimation only.
Recently released large-scale image datasets such as MPIIGaze~\cite{Zhang2015CVPR} and GazeCapture~\cite{Krafka2016CVPR} enable so-called cross-person (person-independent) gaze estimation, where models are evaluated on data from people that were unseen during training.
Various CNN architectures have since been proposed to improve gaze estimation~\cite{cheng2018appearance,park2018deep,Zhang2017CVPRW,wang2019generalizing}, demonstrating their efficacy on additional in-the-wild~\cite{kellnhofer2019gaze360}, laboratory~\cite{fischer2018rt,park2020towards,zhang2020eth,funes2014eyediap}, and synthetic~\cite{wood2016learning,Wood2015ICCV} datasets.

Most of these approaches take as input either cropped eye or face images~\cite{Krafka2016CVPR, park2018deep,cheng2018appearance,fischer2018rt,kellnhofer2019gaze360} or so-called ``data normalized'' images~\cite{Park2019ICCV,Zhang2017CVPRW,zhang2020eth,kothari2021weakly} and do not directly learn to predict gaze from the full camera frame.
The GazeOnce method utilizes multi-task learning to output face existence, face location, facial landmarks, and 3D gaze direction with the raw frame as the input \cite{zhang2022gazeonce}, however, this method 
does not output the gaze origin nor allow for eventual PoG estimation.
While few-shot adaptation approaches~\cite{Park2019ICCV,linden2019learning,zhang2019evaluation} still have the opportunity to adapt to systematic errors due to the user or technical setup, the performance of most cross-person methods will vary greatly depending on the pre-processing adopted during inference time.
Our paper demonstrates that it is possible to learn the complex mapping between frames and 6D gaze rays by designing appropriate modules for the sub-tasks of origin and direction regression.
As we learn to side-step complex pre-processing such as facial landmark detection or data normalization, the performance of our approach should not depend on technical setup related factors.

\begin{figure*}[t]
  \centering
  \includegraphics[width=0.95\linewidth]{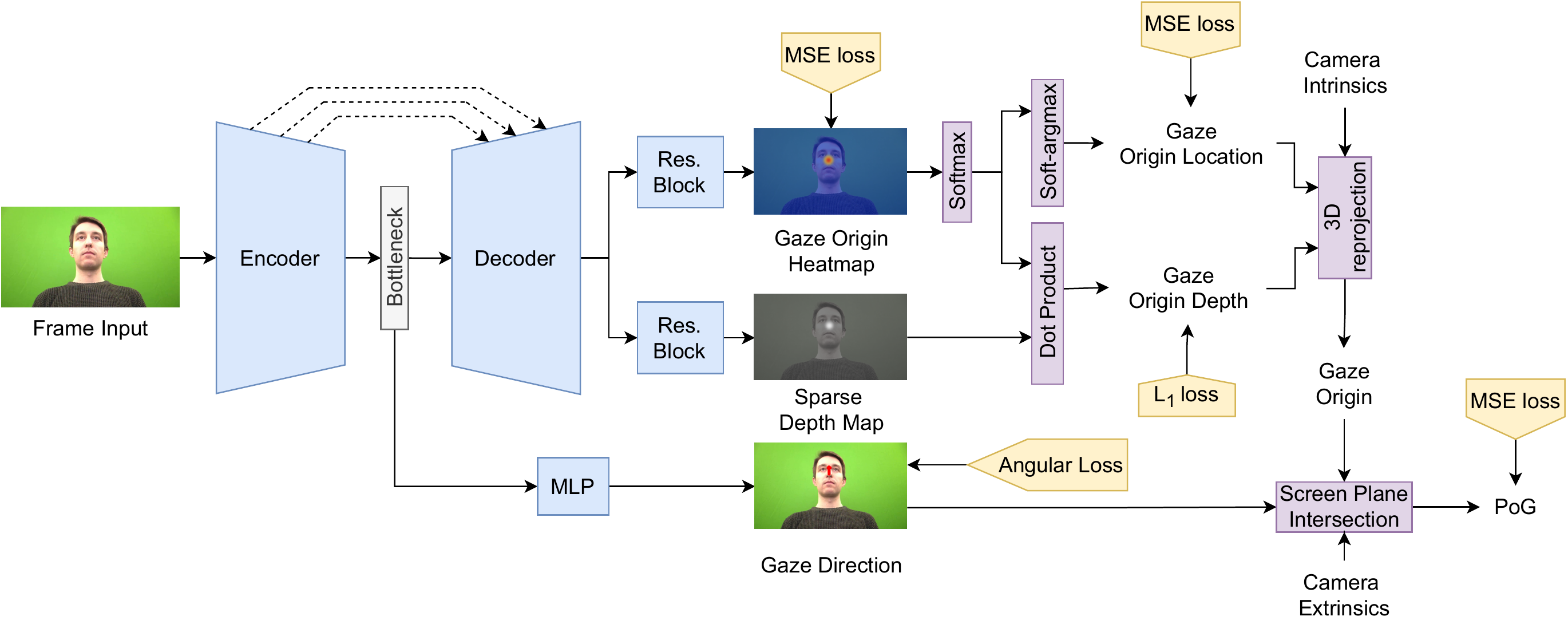}
  \caption{\textbf{The proposed end-to-end frame-to-gaze estimation architecture, \methodname.} We propose a U-Net-like architecture where the output features are mapped to the 2D gaze origin location on the image and a sparse depth map, which are combined to produce the 3D gaze origin. The 3D gaze direction is predicted with an MLP using the bottleneck features as input. The PoG is calculated using predicted gaze origin and direction, together with camera transformation matrices (that define camera-to-screen geometry).}
  \label{fig:overview}
\end{figure*}

\subsection{Learning-based PoG Estimation}
Of the learning-based gaze estimation methods that take RGB input, very few study the task of directly predicting the Point-of-Gaze (PoG).
For example, \cite{Krafka2016CVPR} assumes that all PoGs are on the $z$-plane of the camera coordinate system and directly regress PoG in centimeters, while \cite{Zhang2017CVPRW,huang2017tabletgaze} directly regresses PoG (in cm or mm) regardless of changes in camera-to-screen geometry (rotation, translation, and scaling) between dataset participants.
Other methods for estimating PoG take advantage of the displayed stimuli and evaluate their saliency~\cite{sugano2012appearance} or visual features~\cite{park2020towards} to correct the errors in estimated PoG.
In \cite{park2020towards}, estimated gaze direction is more explicitly composed with known pseudo-ground-truth gaze origin (which is produced as a result of ``data normalization'') and this is combined with known camera-to-screen geometry to produce PoG.
We follow this explicit geometric decomposition in our work but propose to predict gaze origin via a neural network, removing the need for ``data normalization'' at both training and inference times.

\subsection{End-to-end Learning in Gaze Estimation}
In the field of gaze estimation, only very few works extend their methods beyond the eye/face patch input or gaze direction output.
\cite{Krafka2016CVPR,Zhang2017CVPRW}~propose architectures that take eye/face crops as input and yield PoG.
\cite{zhang2022gazeonce}~proposes a multi-person gaze direction estimation method that begins from a large input image, but their method does not predict gaze origin and thus PoG cannot be directly computed.
Our method, \methodname, begins from camera frames and ends with PoG and can be trained end-to-end.
As we learn the relation between camera output and final quantity of interest (PoG), we could consider our method to be truly end-to-end.
\section{Method} \label{sec:method}
Our aim is to learn a model that can estimate a 6D gaze ray, including a gaze origin and a gaze direction, directly from a camera frame in an end-to-end fashion. We call our method End-to-end Frame-to-gaze Estimation (EFE).
To achieve this goal, we decompose the task into two by predicting a gaze origin and a corresponding gaze direction. 
We denote the input RGB image as $\mathbf{X}\in \mathbb{R}^{3\times H \times W}$ pixels, and define the 6D gaze ray as consisting of the gaze origin $\mathbf{o}\in \mathbb{R}^3$ and the gaze direction $\mathbf{r}\in \mathbb{R}^3$. Given camera parameters represented by the intrinsic camera matrix $\mathbf{K}$ and the extrinsic matrix $\mathbf{T}$, we compute the PoG $\mathbf{p}\in \mathbb{R}^3$ on screen.
An overview of EFE is shown in Fig. \ref{fig:overview}.

\subsection{Predicting Gaze Origin}
\label{subsec:origin}

A straightforward approach to predict the gaze origin $\mathbf{o}\in \mathbb{R}^3$ is directly regressing the 3D location coordinates. However, estimating depth through a monocular RGB image is ill-posed and challenging.
Instead of regressing the 3D location, it is well-known that predicting the heatmaps could generate a better estimation,
as shown in the areas of facial landmark localization~\cite{Bulat_2017_ICCV} and human pose estimation~\cite{tompson2014joint}. 
Motivated by that, we propose a U-Net-like architecture to predict a 2D gaze origin heatmap $\mathbf{h} \in \mathbb{R}^{H\times W}$ and a sparse depth map $\mathbf{d}\in \mathbb{R}^{H\times W}$ (see Fig. \ref{fig:overview} for a visual example). 

The positions of gaze origins are not well defined. Most datasets' ground truth labels are created through facial landmark detection and data normalization~\cite{Sugano2014CVPR}. Therefore, learning a distribution of the prediction is more appropriate than predicting a single point since the ground truth labels are likely to contain some degree of error. 
We use $\mathbf{h}$ to predict the 2D location of gaze origin $g \in \mathbb{R}^2$ on the camera frame. We use the mean squared error loss $\mathcal{L}_{\textbf{heatmap}}$ for heatmap prediction and the final $\mathbf{h}$ is obtained after soft-argmax operation~\cite{yi2016lift,chapelle2010gradient}. %

\begin{equation}
\mathcal{L}_{\textbf{heatmap}} = \frac{1}{n}\sum_{i=1}^n\lVert\mathbf{h}-\hat{\mathbf{h}}\rVert^2_2,
\end{equation}
where $n=H \times W$, and $\hat{\mathbf{h}}$ is the ground truth heatmap generated by drawing a 2D Gaussian centered at the gaze origin. The predicted $g$ location is similarly supervised by a mean squared error loss $\mathcal{L}_{g}$,
\begin{equation}
\mathcal{L}_{\mathbf{g}} = \lVert\mathbf{g}-\hat{\mathbf{g}}\rVert^2_2,
\end{equation}
where the $\hat{\mathbf{g}}$ is the ground truth 2D gaze location on the camera frame. The dot product $\mathbf{h} \cdot \mathbf{d}$ is used to predict the gaze depth $z \in \mathbb{R}$. Note that we do not use any approximated depth map for supervision and that the $\mathbf{d}$ is solely learned from $\mathcal{L}_d$ which is defined as 
\begin{equation}
    \mathcal{L}_d = \lVert \mathbf{z} - \hat{\mathbf{z}}\rVert_1,
\end{equation}
where $\hat{\mathbf{z}}$ is the ground truth depth value. 
Our experiments show that this approach outperforms a baseline of regressing the 3D location in the coordinates. 
The gaze origin $\mathbf{o}\in \mathbb{R}^3$ is then calculated by transforming the image coordinates to world coordinates utilizing the camera intrinsic matrix $\mathbf{K}$. Note that $\mathbf{o}_z = \mathbf{z}$.

\subsection{Predicting Gaze Direction}
Since predicting the gaze direction is mapping from image space to the 3D direction, it is not necessary to use heatmap for the estimation. 
As shown in Fig. \ref{fig:overview}, we use the bottleneck features in the middle of the U-Net-like architecture for the prediction of gaze direction $\mathbf{r}\in \mathbb{R}^3$ supervised by the angular loss $\mathcal{L}_{\mathbf{r}}$. The gaze origin and gaze direction prediction share the encoder that could benefit both tasks. The gaze vectors are predicted as Euler angles in spherical coordinates and transformed to 3-dimensional unit vector $\mathbf{r}$. Given the predicted gaze vector $\mathbf{r}$ and ground-truth gaze vector $\hat{\mathbf{r}}$ the angular loss is calculated as
\begin{equation}
    \mathcal{L}_\mathbf{r} = \text{arccos} \left(  \frac{\hat{\mathbf{r}} \cdot  \mathbf{r}} {\lVert \hat{\mathbf{r}} \rVert \lVert  \mathbf{r} \rVert} \right).
\end{equation}

\subsection{Computing PoG}
The PoG is defined by intersecting the 6D gaze ray (composed of origin and direction) with a pre-defined screen plane. 
The screen plane is defined based on the physical screen with its z-axis pointing outwards and towards the user. The estimated PoG should have a z-coordinate of zero in the screen coordinate system, i.e. $\mathbf{p}_z=0$. Using the gaze origin $\mathbf{o}$ and gaze direction $\mathbf{r}$, we can obtain the distance to screen frame $\lambda$. Denoting screen frame normal as $\mathbf{n}_{\text{s}}$ and a sample point on the screen plane as $\mathbf{a}_{\text{s}}$ (e.g. the origin point [0, 0, 0]), we can calculate $\lambda$ as follows:
\begin{equation} \label{eq2}
\lambda = \frac{\mathbf{r} \cdot \mathbf{n}_{s}}{(\mathbf{a}_{s}-\mathbf{o})\cdot \mathbf{n}_{s}}.
\end{equation}
After the distance to the screen frame is calculated, we can find the intersection of the line of sight with the screen plane to compute the PoG $\mathbf{p}$ as follows:
\begin{equation} \label{pog-eq}
\mathbf{p}= \mathbf{o}+ \lambda \mathbf{r}.
\end{equation}
During training, we use mean squared error to supervise PoG estimation:
\begin{equation}
\mathcal{L}_{\text{PoG}} = \lVert\mathbf{p}-\hat{\mathbf{p}}\rVert^2_2
\end{equation}
where the $\hat{\mathbf{p}}$ is the ground truth PoG.

\subsection{End-to-end Frame-to-Gaze Estimation (EFE)}
As shown in Fig. \ref{fig:overview}, the overall \methodname takes a camera image (frame) as input and yields a gaze origin heatmap $\mathbf{h}$ and a sparse depth map $\mathbf{d}$ through two separate residual blocks. Using the intrinsic camera matrix, the origin $\mathbf{o}$ in 3D space is calculated. The gaze direction $\mathbf{r}$ is predicted using the bottleneck features. Lastly, using $\mathbf{o}$ and $\mathbf{d}$, a 6D gaze ray is formed and intersected with a given screen plane to compute PoG. The complete loss is
\begin{equation}
  \mathcal{L_{\textbf{Total}}} = \lambda_g \mathcal{L}_{\mathbf{g}} + \lambda_h \mathcal{L}_{\textbf{heatmap}} + \lambda_d\mathcal{L}_{\mathbf{d}} + \lambda_r\mathcal{L}_{\mathbf{r}} + \lambda_{PoG}\mathcal{L}_{\textbf{PoG}},
\end{equation}
where the first three losses $\mathcal{L}_{\mathbf{g}}$, $\mathcal{L}_{\textbf{heatmap}}$ and $\mathcal{L}_{\mathbf{d}}$ are for gaze origin estimation.

\begin{figure}[t]
  \centering
  \begin{subfigure}[m]{0.58\columnwidth}
    \includegraphics[width=\columnwidth]{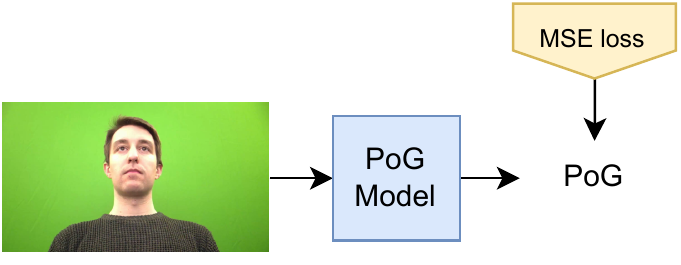}
    \caption{Direct Regression}
  \end{subfigure}
  \\[2mm]
  \begin{subfigure}[m]{1.0\columnwidth}
    \includegraphics[width=\columnwidth]{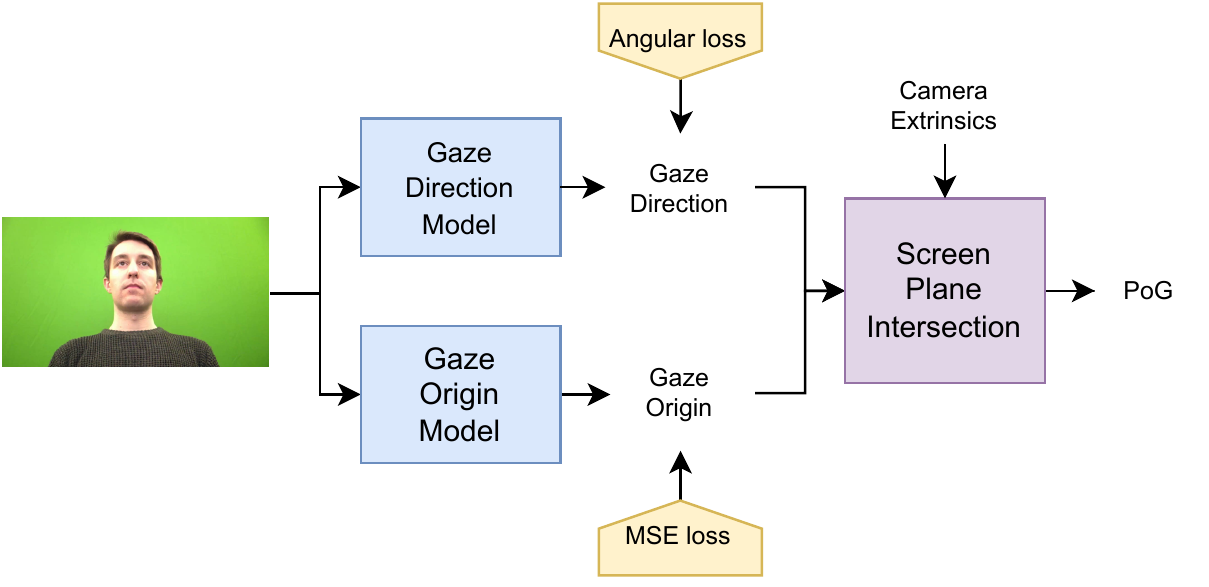}
    \caption{Separate Models}
  \end{subfigure}
  \\[2mm]
  \begin{subfigure}[m]{1.0\columnwidth}
    \includegraphics[width=\columnwidth]{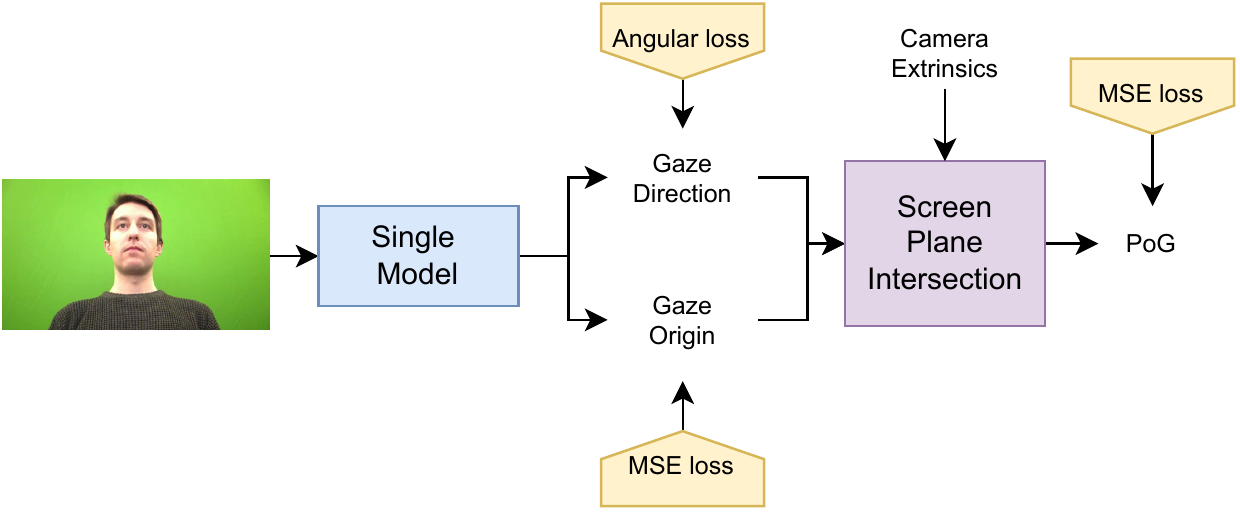}
    \caption{Joint Prediction}
  \end{subfigure}
  \caption{\textbf{Three end-to-end frame-to-gaze baseline models.} (a) Direct Regression: A model that directly estimates PoG, (b) Separate Models: Two separate models estimate gaze origin and gaze direction independently, and (c) Joint Prediction: A single model that jointly predicts both gaze origin and direction. 
  } 
  \label{fig:baselines}
\end{figure}

\section{Experiment}
To demonstrate the effectiveness of \methodname, we first compare it with three end-to-end gaze estimation baselines to show the advantage of our proposed method. %
We then conduct evaluations on three datasets, EVE~\cite{park2020towards}, GazeCapture~\cite{Krafka2016CVPR}, and MPIIFaceGaze~\cite{Zhang2017CVPRW} to show that learning the mapping from frame to gaze is possible and that \methodname does it in a competitive manner when compared to existing state-of-the-art methods.

\subsection{Datasets and Preprocessing}
\parahead{EVE~\cite{park2020towards}}
The EVE dataset is proposed for the end-to-end gaze estimation task. It consists of continuous videos collected from four cameras and uses a Tobii Pro Spectrum (150Hz) eye tracker to provide the ground truth gaze labels. 
Although the EVE dataset is proposed as a video dataset, we create an image dataset using its frames with a subsampling rate of 0.6
for ease of experimentation (only applied to the training set). The frames are resized to $480 \times 270$ pixels in size before using as input to the evaluated models.

\parahead{GazeCapture~\cite{Krafka2016CVPR}} The GazeCapture dataset is collected through crowdsourcing with phones and tablets. It consists of over 1450 people and almost $2.5M$ frames. The dataset provides the input raw frames and the PoG with respect to the camera while assuming that the camera and screen are coplanar.

\parahead{MPIIFaceGaze~\cite{Zhang2019TPAMI}} The MPIIFaceGaze dataset is collected from 15 subjects with their laptop under natural head movements and diverse lighting conditions. There are 3000 face images for each subject. It provides both raw camera frames, the 3D gaze origins estimated by data normalization, the 3D gaze directions, and the 2D PoGs on the screen.

\subsection{Implementation Details}
Our U-Net-like architecture takes the EfficientNet-V2 small \cite{EfficientNetV2} architecture as the backbone. %
We observe that $\mathcal{L}_{\textbf{PoG}}$ is significantly higher than the other losses in the early stages of training, as the model has not yet been able to estimate the gaze origin and direction with reasonable accuracy. Therefore, we do not optimize with respect to $\mathcal{L}_{\textbf{PoG}}$ for the first two epochs of the training process and set $\lambda_{PoG}=0$. %
We train \methodname using the AdamW optimizer \cite{adamW} for eight epochs unless otherwise mentioned, using a batch size of 32. An exponential learning rate decay of factor 0.9 is applied, beginning from a learning rate of 0.0003. The origin and PoG coordinates are standardized with train set metrics for each of the datasets. $\lambda_g$ is set to 2 and the remaining loss weights are set to 1.

\begin{table*}
  \centering
  \begin{tabular}{lccccc}
    \toprule
    Model & Heatmap pred. & Depth map pred. & Gaze Origin (mm) & Gaze Dir. (\textdegree) & PoG (px)\\
    \midrule
    Direct Regression & & &  - & - & 143.83 \\
    Separate Models & & & 16.18 & 3.63 & 141.35 \\
    Joint Prediction & & & 20.14 & 3.73 & 143.39 \\
    \methodname w/o depth map & \checkmark & & 18.28 & 3.82 & 146.82 \\
    \methodname~(ours) & \checkmark & \checkmark & \textbf{16.07} & \textbf{3.53} & 
    \textbf{133.73}\\
    \bottomrule
\end{tabular}
  \caption{\textbf{Comparison to the baseline models on the EVE dataset.} Accuracy of estimated gaze origin is reported in millimeters (mm), gaze direction in degree (\textdegree), and PoG in screen pixels (px). The first three baselines correspond to the three models shown in Fig. \ref{fig:baselines}. Note that the input frames are the raw frames from the camera without any face and facial landmark detection.}
\label{tab:baselines}
\end{table*}

For the EVE dataset, we terminate the training after 8 epochs and set $\lambda_{PoG}=0$ in the first 2 epochs. The inputs to \methodname are $480\times 270$ pixels in size %
resized from the original $1920\times 1080$ pixels in the dataset.

For GazeCapture, we terminate the training after five epochs and set $\lambda_{PoG}=0$ for the first epoch as it is a large dataset. The PoG prediction is truncated to the screen size similar to the original work using the device type and orientation information. The inputs to \methodname are $512\times 512$ raw camera frames created from the original dataset\footnote{The GazeCapture dataset originally consists of images of size $640\times 480$ and $480 \times 640$ due to the diverse mobile device orientations used.} by re-projecting using a common virtual camera with a focal length of $460$ mm to consolidate images from diverse camera devices and orientations.

For MPIIFaceGaze, we terminate training after 15 epochs and set $\lambda_{PoG}=0$ in the first three epochs. The inputs to \methodname are $640\times 480$ raw camera frame resized from $1280\times 720$ pixels in the original dataset, and are created by re-projecting using a common virtual camera with a focal length of $550$ mm.

\subsection{End-to-End Frame-to-Gaze Baseline Models}

Given the raw frame from the camera, we consider three baseline models: \textit{direction regression}, \textit{separate models}, and \textit{joint prediction} as shown in Fig. \ref{fig:baselines}. The direction regression method directly estimates PoG without predicting any gaze origin or gaze direction. The separate models method estimates the gaze origin and direction independently through two identical models. The joint prediction method estimates gaze origin and direction via a shared convolutional neural network and two separate MLPs. Importantly, the joint prediction method is optimized with a PoG loss as well which should improve PoG estimation performance. The joint prediction baseline is the most similar one to \methodname. However, the main difference between \methodname and these baselines stems from the gaze origin prediction architecture. The heatmap-based prediction of gaze origin allows \methodname to model the uncertainty, and the prediction of origin depth through sparse depth maps allows this uncertainty to be propagated to the depth prediction.

We evaluate the three baselines and \methodname on the EVE dataset since the dataset is specifically designed for the end-to-end gaze estimation task.
Also, EVE uses a complex and offline data pre-processing scheme (including per-subject calibration of a morphable 3D face model) and thus the gaze origin labels acquired via data normalization are of high quality, making the dataset a particularly challenging benchmark for Frame-to-Gaze methods.
We show in Tab. \ref{tab:baselines} that \methodname outperforms all three baselines. The direct regression neither predicts gaze origin nor gaze direction and achieves poor performance compared to the other methods. The separate models method achieves better results than the other two baselines due to the larger model capacity. %
The joint prediction is similar to our architecture and achieves worse results than the separate models, which indicates that it cannot learn gaze origin and direction jointly in an effective manner. 
In contrast, \methodname achieves the best performance compared with all three baselines. %

Furthermore, we can see from Tab. \ref{tab:baselines} that performance degrades greatly when we do not use the sparse depth map estimation module, instead, using a MLP to predict $\mathbf{z}$ from bottleneck features.
This demonstrates that predicting a heatmap for gaze origin regression combined with the learning of a sparse depth map for distance estimation is important for Frame-to-Gaze architectures.

\subsection{Comparison with State-of-the-art}
In this section, we compare \methodname with state-of-the-art methods on three datasets, i.e. EVE, GazeCapture, and MPIIFaceGaze.

\begin{table}
  \begin{tabular}{lc
  >{\centering\arraybackslash}p{15mm}
  >{\centering\arraybackslash}p{10mm}}
    \toprule
    Model & Inputs & Gaze Dir. (\textdegree) & PoG (px)\\
    \midrule
    EyeNet (static)~\cite{park2020towards} & Right Eye & 4.75 & 181.0 \\
    EyeNet (static)~\cite{park2020towards} & Left Eye & 4.54 & 172.7 \\
    FaceNet & Face & \textbf{3.47} & 134.10  \\
    \methodname (ours) & Frame & 3.53 & \textbf{133.73} \\
    \bottomrule
  \end{tabular}
  \caption{\textbf{Comparison with state-of-the-art methods on the EVE dataset.} Accuracy of estimated gaze direction is reported in degree (\textdegree), and PoG in screen pixels (px).
  }
\label{tab:norm}
\end{table}

\parahead{Comparison with SotA on EVE}
For the EVE dataset, we list the performance of EyeNet reported in the original EVE paper \cite{park2020towards}. EyeNet uses either left or right-eye images as input, and predicts gaze direction and PoG independently for each eye. Note that EyeNet uses ground-truth gaze origins acquired via data normalization, similar to most other state-of-the-art baseline methods.
In addition, we train a model with the face images acquired from the data normalization procedure and denote it \textit{FaceNet}. It uses the ground truth gaze origin acquired through data normalization and the model itself only outputs gaze direction. This is the most common and high-performing problem formulation in learning-based gaze estimation. As in \methodname, we use the same EfficientNet-V2 small \cite{EfficientNetV2} as the backbone of FaceNet for a fair comparison.

As shown in Tab. \ref{tab:norm}, EyeNet achieves the worst results even with the normalized eye images. By taking a normalized face image as input, FaceNet achieves better results, in line with the literature on face-based gaze estimation~\cite{Krafka2016CVPR,Zhang2017CVPRW}. The input to FaceNet is a normalized face image of size $256\times 256$, while the input to \methodname is a resized raw frame of size $480 \times 270$. Note that the effective face resolution is much smaller in the resized frame than in the normalized face image. Nonetheless, \methodname has comparable performance to FaceNet.
Thanks to end-to-end learning with a direct PoG loss, we find that \methodname has a slightly better PoG estimate despite having worse performance on gaze direction.
This could be attributed to the fact that the ground-truth gaze labels provided in EVE are themselves estimates and contain some degree of error. 
We expect higher accuracy of the provided ground truth PoG as it is measured by a desktop eye tracker, and a direct loss using this PoG ground-truth would result in better learning and consequent model performance.
Indeed, our \methodname performs best in terms of PoG prediction. 

\vfill\null\columnbreak

\begin{table}
  \centering
  \begin{tabular}{lc
  >{\centering\arraybackslash}p{9mm}
  >{\centering\arraybackslash}p{10mm}
  }
    \toprule
    Model & Inputs & Phone PoG & Tablet PoG\\
    \midrule
    iTracker~\cite{Krafka2016CVPR} & Face\&Eyes & 2.04 & 3.32 \\
    iTracker {\small(train aug)}~\cite{Krafka2016CVPR} & Face\&Eyes & 1.86 & 2.81 \\
    SAGE~\cite{he2019device} & Eyes & 1.78 & 2.72 \\
    TAT~\cite{guo2019generalized} & Face & 1.77 & 2.66 \\
    AFF-Net~\cite{bao2021adaptive} & Face\&Eyes & \underline{1.62} & \textbf{2.30} \\
    \methodname (ours) & Frame & \textbf{1.61} &  \underline{2.48} \\
  \bottomrule
\end{tabular}
  \caption{\textbf{Comparison with state-of-the-art methods on the GazeCapture dataset.} Accuracy of estimated PoG is reported in centimeters (cm).
   \label{tab:gazecap}}
\end{table}

\parahead{Comparison with SotA on GazeCapture}
We show the comparison between \methodname and other state-of-the-art on the GazeCapture dataset in Tab. \ref{tab:gazecap}.  
The results of iTracker~\cite{Krafka2016CVPR} are reported from the original paper and we list performances for both \textit{phone} and \textit{tablet} on the GazeCapture dataset. The iTracker method takes multiple inputs: the cropped face, left eye, right eye, and face occupancy grid. In contrast, \methodname takes only the resized raw frame as input. From the table, we can see that \methodname outperforms iTracker by a large margin. It shows that the proposed end-to-end method is better than using multiple cropped images for direct PoG regression.

In comparison with more advanced methods such as a method using knowledge distillation~\cite{guo2019generalized} and a method taking eye corner landmarks as input~\cite{he2019device}, our \methodname still performs competitively, with lower PoG error on both \textit{phone} and \textit{tablet} subsets.
Note that \methodname only requires the full frame as input (albeit re-projected to a common set of camera parameters) while all other methods require multiple specifically prepared inputs such as cropped face image \cite{Krafka2016CVPR,guo2019generalized}, cropped eye image \cite{Krafka2016CVPR,he2019device}, ``face grid'' image \cite{Krafka2016CVPR}, and/or eye corner landmarks \cite{Krafka2016CVPR,he2019device}.
\methodname achieves comparable performance to the latest PoG estimation method AFF-Net~\cite{bao2021adaptive}, which not only takes the cropped face and eye patches as input but also incorporates complex feature fusion between these patches.

\begin{table} 
  \centering
  \begin{tabular}{lc
  >{\centering\arraybackslash}p{14mm}
  >{\centering\arraybackslash}p{10mm}}
    \toprule
    Model & Input & Gaze Dir. (\textdegree) & PoG (mm)\\
    \midrule
    Full-Face~\cite{Zhang2017CVPRW} & Face & 4.8 & \underline{42.0} \\
    FAR-Net*~\cite{cheng_2020} &Face\&Eyes& \textbf{4.3} & - \\
    AFF-Net~\cite{bao2021adaptive} &Face\&Eyes& \underline{4.4} & 39.0 \\
    EFE (ours) & Frame & \underline{4.4} & \textbf{38.9}  \\
  \bottomrule
\end{tabular}
  \caption{\textbf{Comparison with state-of-the-art methods on the MPIIFaceGaze dataset.} 
  Accuracy of estimated gaze direction is reported in degrees (\textdegree), and PoG in millimeters (mm).
  }
  \label{tab:mpii}
\end{table}

\vspace*{4mm}

\parahead{Comparison with SotA on MPIIFaceGaze}
For the MPIIFaceGaze dataset, we compare \methodname with state-of-the-art PoG estimation methods, Full-Face~\cite{Zhang2017CVPRW}, FAR-Net*~\cite{cheng_2020}, and AFF-Net~\cite{bao2021adaptive}, under a 15-fold cross-validation evaluation scheme. The numbers are copied from the corresponding papers.
As seen in Tab. \ref{tab:mpii}, \methodname outperforms the Full-Face approach and is comparable to the FAR-Net* approach and AFF-Net.
Note that Full-Face,  FAR-Net*, and AFF-Net use face images after data normalization as input as well as the ground truth gaze origin computed via data normalization, therefore, the model itself only outputs the gaze direction. Although the proposed \methodname takes as input the raw frame, which has fewer effective pixels on the face region, it still achieves comparable results. Once again, the results show that the proposed \methodname gaze estimation pipeline is effective even without the complex data normalization procedure.

\subsection{Qualitative Analysis}

\begin{figure}[t]
  \vskip -4mm
  \centering
  \hfill
  \begin{subfigure}[t]{0.48\columnwidth}\centering
    \includegraphics[width=\columnwidth,trim={5mm 5mm 5mm 5mm},clip]{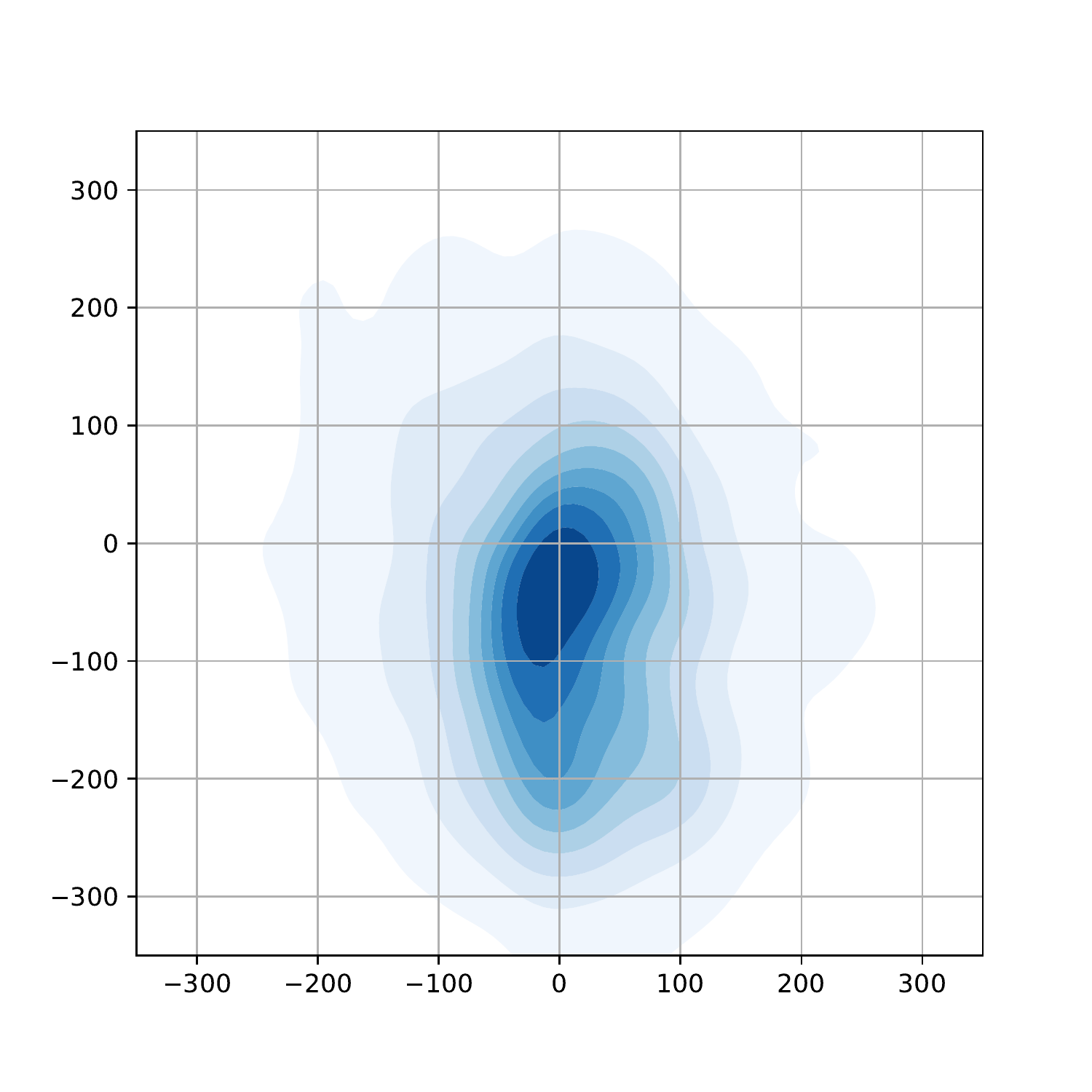}
  \end{subfigure}
  \hfill
  \begin{subfigure}[t]{0.48\columnwidth}\centering
    \includegraphics[width=\columnwidth,trim={5mm 5mm 5mm 5mm},clip]{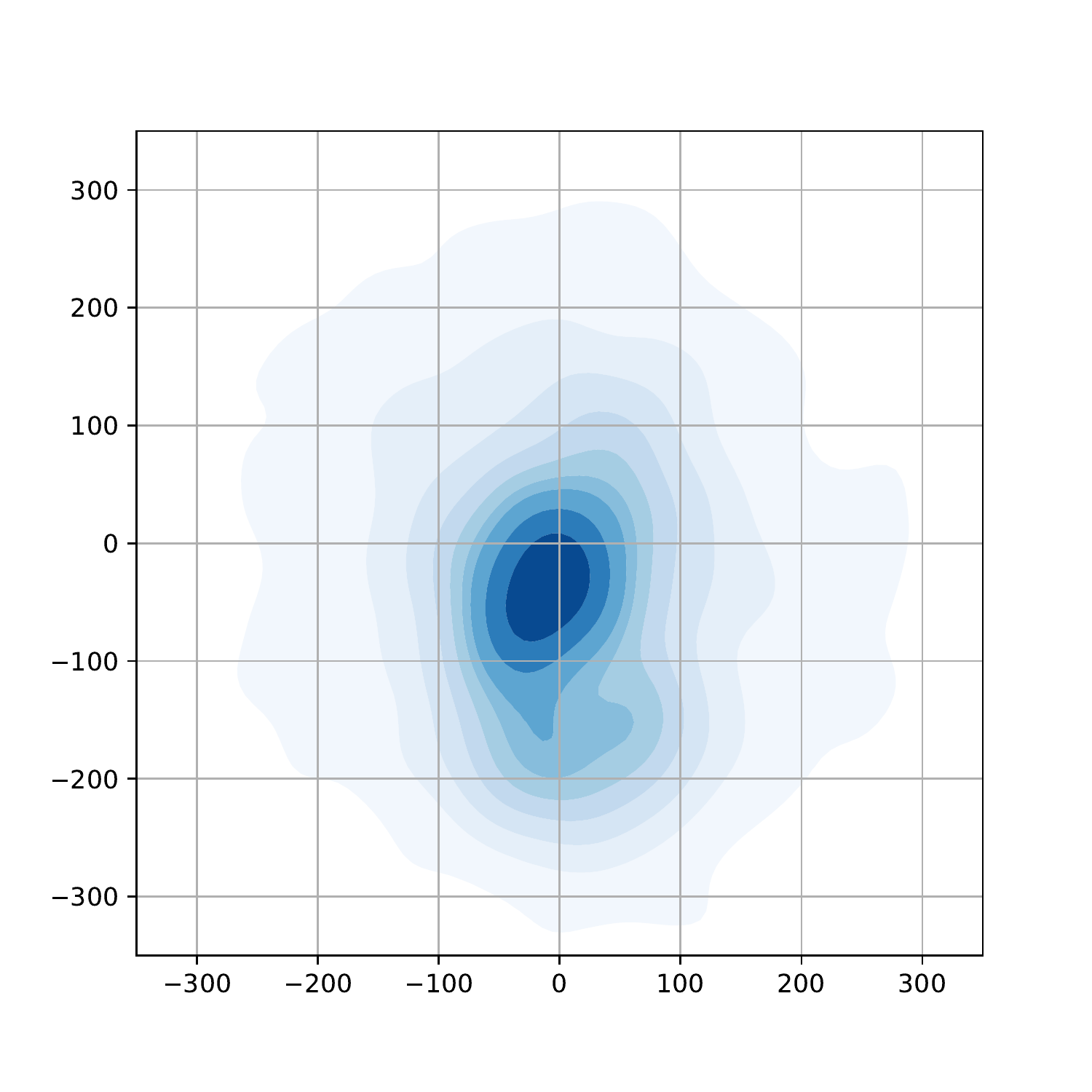}
  \end{subfigure}
  \vskip -3mm
  \caption{\textbf{2D histogram of PoG residuals on EVE.} PoG error distribution on the screen in pixels with the same predicted gaze direction but different gaze origins. \textbf{Left:} gaze origin is calculated by the data normalization. \textbf{Right:} gaze origin is predicted by \methodname. The comparison shows that our end-to-end learning approach exhibits less bias in its PoG errors.
  \label{fig:pog_residuals}
  }
  \vskip -9mm
\end{figure}

In this section, we analyze our proposed method, \methodname, in a qualitative manner.

\parahead{Bias in PoG residuals}
The majority of gaze direction estimation methods use gaze origin determined via the data normalization procedure as ground truth.  
Blindly relying on this gaze origin value can result in systematic errors.
In contrast, \methodname trains with a direct PoG loss which is more reliable and can correct the systematic errors caused by data normalization.
We observe this in Fig. \ref{fig:pog_residuals} where we find that \methodname shows lower bias in terms of PoG residuals, in that its error distribution is more centered and isotropic.

\begin{figure*}[t]
  \includegraphics[width=\textwidth]{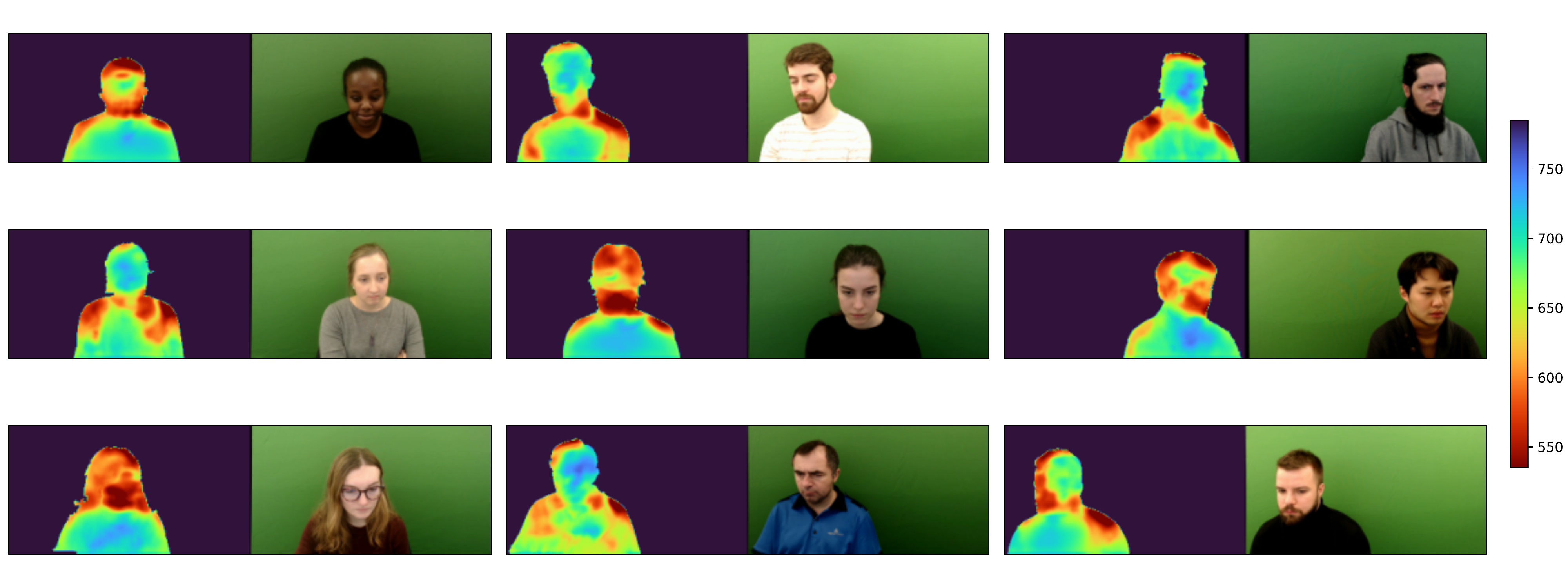}
  \vskip -4mm
  \caption{\textbf{Depth maps predicted by \methodname.} The depth maps are colored using the turbo colormap~\cite{mikhailov2019turbo} and the green background is subtracted for EVE to make the visualizations easier to interpret. Though not all depth values match real-world expectations, the model has an approximate notion of whether the head and torso are further away or closer to the camera. 
  \label{fig:depthmaps}
  }
  \vspace*{-3mm}
\end{figure*}

\parahead{Visualization of Predicted Depth Maps}
The depth map predicted by \methodname is supervised via a point-like loss ($\mathcal{L}_\mathbf{d}$), which only refers approximately to the face region of the visible user.
Yet surprisingly, Fig. \ref{fig:depthmaps} shows that somewhat plausible depth maps can be predicted.
More specifically, a rough notion of whether the user's face is further away or closer to the camera is captured.
It is interesting that such weak supervision can still produce plausible dense outputs.

\subsection{Cross-camera Evaluation}
The data normalization procedure is introduced with the motivation that it produces cropped patches that are more agnostic to the specific camera used.
Consequently, the models trained using cropped patches would generalize across cameras.
To evaluate the generalization across cameras, we conduct a cross-camera evaluation on the EVE dataset with its four cameras, comparing \methodname and the data normalization-based approach FaceNet. 

We show the result in Tab. \ref{tab:cross_camera}.
From the table, we find that \methodname generalizes surprisingly well to extreme camera view changes, exhibiting large performance differences to a data normalization-based approach (FaceNet).
For example, \methodname achieves 13.2$^\circ$ gaze direction error when training on the \emph{W$_L$} camera and testing on the \emph{W$_R$} camera, which is a 16.8$^\circ$ improvement compared to FaceNet.
This is likely due to the way in which \methodname decomposes the PoG estimation problem into the two sub-tasks of gaze origin and gaze direction estimation.

\begin{table}[t]
    \centering
    \begin{subtable}[h]{0.48\textwidth}
        \centering
        \begin{tabular}{l|cccc}
        \toprule
        Train / Test    & MVC    & W$_C$ & W$_L$ & W$_R$ \\
        \midrule
        MVC             &    -    & 31.6 & 36.8        & 42.0 \\
        W$_C$ & 30.6 &       -         & 10.9        & 11.6 \\
        W$_L$ & 31.3 & \textbf{6.9} &    -        & 30.0 \\
        W$_R$ & 33.1 & \textbf{7.3} & 22.1 &  -    \\
        \bottomrule
        \end{tabular}
        \caption{Errors achieved by FaceNet}
    \end{subtable}
    \\[2mm]
    \begin{subtable}[h]{0.48\textwidth}
        \centering
        \begin{tabular}{l|cccc}
        \toprule
        Train / Test    & MVC    & W$_C$ & W$_L$ & W$_R$ \\
        \midrule
        MVC   & - & \textbf{23.4} & \textbf{24.8} & \textbf{28.4} \\
        W$_C$ & \textbf{23.0} & - & \textbf{10.1} & \textbf{11.0} \\
        W$_L$ & \textbf{23.3} & 11.1 & - & \textbf{13.2} \\
        W$_R$ & \textbf{28.0} & 11.8 & \textbf{14.5} & - \\
        \bottomrule
        \end{tabular}
        \caption{Errors achieved by \methodname}
    \end{subtable}
    \\[-2mm]
    \caption{
        \textbf{Cross-camera gaze direction error ($^\circ$).}
        We conduct a challenging cross-camera and cross-person evaluation where models are only trained with one camera view and tested on another camera view. We use four cameras in the EVE dataset including the machine vision camera (MVC) under the display, the webcam on the top (W$_C$), top left (W$_L$), and top right (W$_R$) of the display.
        We see that \methodname outperforms the data normalization-based FaceNet in most cross-camera configurations, by large margins in many cases.
        \label{tab:cross_camera}
    }
    \vskip -2mm
\end{table}
\section{Conclusion}
Existing methods in gaze estimation have typically relied on simple cropping \cite{Krafka2016CVPR} or data normalization~\cite{Sugano2014CVPR,Zhang2018ETRA} as a scheme for simplifying the gaze direction estimation problem and for achieving higher performances.
In this paper, we challenged this paradigm and proposed 
an architecture that can predict gaze origin from input camera frames directly.
To the best of our knowledge, we are the first to demonstrate a solution to this problem in a learning-based approach and end-to-end manner.
Furthermore, we were able to show that despite the smaller effective face size in the frame-to-gaze setting, the proposed \methodname is able to achieve comparable performances to state-of-the-art methods.
Future work could expand our proof-of-concept by proposing alternative methods of breaking down the complex problem of PoG estimation on mobile and edge devices in many practical settings.

\parahead{Limitations}
The camera-to-screen geometry must be known \emph{a priori} for \methodname~-- a limitation that we share with data normalization-based works.
Few-shot learning methods could be adopted in the future to adapt to novel camera-to-screen geometries with minimal user input.

{\small
\bibliographystyle{ieee_fullname}
\bibliography{main}
}

\end{document}